\documentclass{article}

\usepackage{times}
\usepackage{booktabs}
\usepackage{graphicx} 
\usepackage{subfigure} 

\usepackage{natbib}

\usepackage{algorithm}
\usepackage{algorithmic}

\usepackage{hyperref}

\usepackage{amsmath}	
\usepackage{amsfonts}
\usepackage{wrapfig}
\usepackage{amssymb}
\usepackage{amsthm}
\usepackage{cuted}
\usepackage[arxiv]{icml2015} 
\usepackage[arxiv]{icml2015}

\newcommand{\cnteq}{\nonumber\\}

\renewcommand{\vec}[1]{{\bf #1}}

\newcommand{\hv}{\vec{h}}
\newcommand{\vv}{\vec{v}}

\newcommand{\hi}[1][i]{h_{#1}}
\newcommand{\vj}[1][j]{v_{#1}}

\newcommand{\av}{\vec{a}}
\newcommand{\bv}{\vec{b}}

\newcommand{\ai}[1][i]{a_{#1}}
\newcommand{\bj}[1][j]{b_{#1}}

\newcommand{\ESS}{\mathrm{ESS}}
\newcommand{\aisw}{{w^{(i)}}}
\newcommand{\naisw}{{w_*^{(i)}}}

\newcommand{\wij}[1][ij]{W_{#1}}

\newcommand{\real}{\mathbb{R}}
\newcommand{\binary}{\{0,1\}}

\newcommand{\dd}{\mathrm{d}}

\newcommand{\refeq}[1]{Eq.~(\ref{#1})}
\newcommand{\reffig}[1]{Fig.~\ref{#1}}
\newcommand{\refFig}[1]{Figure~\ref{#1}}
\newcommand{\refalg}[1]{Algorithm~\ref{#1}}
\newcommand{\refth}[1]{Theorem~\ref{#1}}
\newcommand{\reftab}[1]{Table~\ref{#1}}

\newcommand{\Exp}{\mathbb{E}}

\newcommand{\pstr}{p^*}
\newcommand{\pstrb}[1][\beta]{p^*_{#1}}
\newcommand{\pb}[1][\beta]{p_{#1}}
\newcommand{\obj}{{\cal J}(\beta(\cdot))}

\newcommand{\params}{{\boldsymbol\theta}}
\newcommand{\para}{{\boldsymbol\theta}}

\newcommand{\parab}{{\boldsymbol\theta}}

\newcommand{\erbm}{E}

\newcommand{\vbeta}{\dot{\beta}}
\newcommand{\abeta}{\ddot{\beta}}
\newcommand{\var}{\mathrm{Var}}
\newcommand{\varb}[1][\beta]{{\var_{#1}}}
\newcommand{\tk}[1][k]{t_{#1}}

\newcommand{\wais}{w}

\newcommand{\zB}{Z_\mathrm{B}}
\newcommand{\paraB}{\params^\mathrm{B}}
\newcommand{\pB}{p_\mathrm{B}}

\newcommand{\paraA}{\params^\mathrm{A}}
\newcommand{\pA}{p_\mathrm{A}}

\newcommand{\AIS}{\mathrm{AIS}}
\newcommand{\betaseq}{\{\beta_k\}}
\newcommand{\betat}[1][t]{\beta(#1)}
\newcommand{\hbetaseq}{\{\tilde{\beta}_k\}}

\newtheorem{theorem}{Theorem}
\newcommand{\ze}{\hat{Z}_\mathrm{B}}

\newcommand{\dom}{\cal V}
\newcommand{\tran}[1][\beta_k]{T_{#1}}
\newcommand{\trans}[1][\beta_k]{\{\tran[#1]\}}

\newcommand{\pk}[1][\beta_{k}]{p_{#1}}
\newcommand{\pseq}[1][\beta_{k}]{\{\pk[#1]\}}

\newcommand{\dmax}{{\Delta\beta_\mathrm{max}}}
\newcommand{\dbeta}[1][k]{{\Delta\beta_{#1}}}

\icmltitlerunning{Variational Optimization of Annealing Schedules}

\begin{document} 

\twocolumn[
\icmltitle{Variational Optimization of Annealing Schedules}

\icmlauthor{Taichi Kiwaki}{kiwaki@sat.t.u-tokyo.ac.jp}
\icmladdress{The University of Tokyo, Tokyo, Japan}

\icmlkeywords{Annealed Importance Sampling, Optimal binning, Restricted Boltzmann Machines, Calculus of Variation}

\vskip 0.3in
]

\begin{abstract} 
Annealed importance sampling (AIS) is a common algorithm to estimate partition functions of useful stochastic models. One important problem for obtaining accurate AIS estimates is the selection of an annealing schedule. 
Conventionally, an annealing schedule is often determined heuristically or is simply set as a linearly increasing sequence. 
 In this paper, we propose an algorithm for the optimal schedule by deriving a functional that dominates the AIS estimation error and by numerically minimizing this functional. We experimentally demonstrate that the proposed algorithm mostly outperforms conventional scheduling schemes with large quantization numbers. 
\end{abstract} 

\section{Introduction}

A large number of useful stochastic models are defined using unnormalized probability. 
Exact computation of normalizing constants, or partition functions, of such models is usually intractable. 
This poses a difficulty in comparing different models or training algorithms with respect to the probability that the models assign to validation data. 
Motivated by this problem, extensive research has been made on estimation of partition functions \cite{Gelman:1998ux, Neal:2001wea, Yedidia:2005wl}. 
Annealed importance sampling (AIS) is a common estimation algorithm for partition functions with a nice property that unbiased estimates are obtained \cite{Neal:2001wea, Salakhutdinov:2008eq, Grosse:2013tf}. 

One of the principal problems for achieving accurate estimates is the selection of an annealing path and an annealing schedule \cite{Gelman:1998ux, Grosse:2013tf, Neal:2001wea}. Though mainstream research has been addressed to the selection of an annealing path \cite{Gelman:1998ux, Grosse:2013tf}, little has been done on the selection of an annealing schedule. Some researchers develop heuristic scheduling \cite{Salakhutdinov:2008eq, Desjardins:2013tz}, and others simply use the linear schedule \cite{Salakhutdinov:2009wv, Dauphin:2013um}. \citet{Grosse:2013tf} recently developed a scheduling algorithm but this algorithm failed to make remarkable improvements over the linear schedule. 

In this paper, we propose an alternative scheduling algorithm by formulating the problem as variational minimization of a functional that dominates the variance of estimates. 
We develop a numerical solver for the variational problem and implement an optimization scheme for an annealing schedule. 
We perform experiments on restricted Boltzmann machines (RBMs) and show that the proposed algorithm outperforms conventional scheduling schemes with a large number of quantization.

\section{Models of Interest}

The schemes discussed in this paper cover stochastic models that assign probabilities 
\begin{align}
p(\vv;\para) &= \frac{p^*(\vv;\para)}{Z(\para)},
\end{align}
to states $\vv\in\dom$ where $\para$ are model parameters, and $\pstr$ is the unnormalized probability that can be efficiently evaluated. The main interest of this paper is to estimate the partition function of such models
\begin{align}
 Z(\para) &\triangleq \sum_{\vv\in\dom} p^*(\vv;\para),
\end{align}
which is often intractable.

One example of such models is RBMs. An RBM is a binary Markov random field with a bipartite graph structure that consists of two layers of variables: visible variables representing data $\vv \in \binary^D$, and hidden variables representing latent features $\hv \in \binary^M$ \cite{Hinton:2002wb}. 
The unnormalized probability of an RBM is computed as 
\begin{align}
p^*(\vv;\para) &= \sum_{\hv} \exp(-E(\hv, \vv;\para)), 
\end{align}
where RBM energy function is defined as
\begin{align}
\erbm(\hv, \vv;\parab) = -\sum_{i=1}^{M}\sum_{j=1}^D \vj \wij \hi - \sum_{i=1}^M \ai\hi  - \sum_{j=1}^D\bj\vj,\label{rbm:eq}
\end{align}
and the parametes are $\parab = \{W, \av, \bv\}$. 

\begin{algorithm}[t]
   \caption{AIS: Annealed Importance Sampling}
   \label{AIS:alg}
\begin{algorithmic}
   \STATE {\bfseries Input:} annealing schedule $\betaseq$, number of runs $N$, function $f:\dom\rightarrow\real$
   \STATE Initialize $\wais^{(\cdot)}=1$ 
   \FOR{$k=1$ {\bfseries to} $K$}
   \STATE Sample  $\vv_0^{(\cdot)}$ from $p_{0}(\vv_0^{(\cdot)})$
   \FOR{$i=1$ {\bfseries to} $N$}
   \STATE Update $\wais^{(i)}\leftarrow \wais^{(i)}\frac{\pstr_{\beta_k}(\vv_{k-1}^{(i)})}{\pstr_{\beta_{k-1}}(\vv_{k-1}^{(i)})}$ 
   \STATE Sample $\vv_k^{(i)}\sim\tran(\vv_k^{(i)}|\vv_{k-1}^{(i)})$
   \ENDFOR
   \STATE $\hat{f}_k = \sum_{j}\wais^{(j)}f(\vv_{k-1}^{(j)})/\sum_{j}\wais^{(j)}$
   \ENDFOR
 \STATE Compute $\ze = Z(\paraA) \sum_i^N \wais^{(i)}/N$
 \STATE {\bfseries Output:} $\ze$, $\{\hat{f}_k\}$
\end{algorithmic}
\end{algorithm}

\section{Annealed Importance Sampling}
Suppose that we are estimating the partition function $\zB\triangleq Z(\paraB)$ of an intractable stochastic model $\pB(\vv)\triangleq p(\vv;\paraB)$ with parameters $\paraB$. A possible way to estimate $\zB$ is to use importance sampling (IS) with some tractable distribution $\pA$. By assuming that $\pA(\vv)\neq0 \Leftarrow \pB(\vv)\neq0$, the partition function can be approximated as: $\zB = \int \frac{p^*_B(\vv)}{p_A(\vv)}p_A(\vv)\dd\vv \approx \ze \triangleq \frac{1}{N}\sum_i^N\frac{p^*_B(\vv_i)}{p_A(\vv_i)}$. This Monte Carlo estimate is unbiased if we obtain i.i.d. samples from $\pA$. However, the variance of estimates can generally be large unless $p_A$ is a close approximation of $p_B$, which is not often the case.

Annealed importance sampling (AIS) eases this problem by using a sequence of intermediate distributions $\pseq$ defined with a sequence $0=\beta_0<\ldots<\beta_K=1$ that interpolates between $\pA$ and $\pB$, i.e., $\pk[\beta_0=0]=\pA$ and $\pk[\beta_K=1]=\pB$ \cite{Neal:2001wea, Salakhutdinov:2008eq, SohlDickstein:2012tk}. 
AIS alternates between importance weight updates and annealed MCMC updates as in \refalg{AIS:alg} where $\tran[\beta]$ is an MCMC transition operator that renders $\pk[\beta]$ invariant.
Note that AIS shares the unbiasedness property with IS. Remarkably, unbiasedness holds even if MCMC transitions do not return independent samples \cite{Neal:2001wea}. 

As \citet{Neal:2001wea} suggests, the effective sample size (ESS) can be an informative measure for estimation accuracy of AIS. The ESS can be estimated as:
\begin{align}
 \ESS = \frac{N}{1+s^2(\naisw)},
\end{align}
where $s^2(\naisw)$ is the sample variance of $\naisw = N\aisw/\sum_{i=1}^N \aisw$. The ESS is approximately inversely proportional to the variance of AIS estimates and is reliable unless AIS samples are misallocated to major modes \cite{Neal:2001wea}. 

One interesting property of AIS is that the statistics of intermediate distributions can be estimated with on-the-fly importance weights at any point of annealing \cite{Neal:2001wea}. For example, an expectation of some function $f(\vv)$ with respect to $\pk$ can be estimated as $\hat{f}_k$ as in \refalg{AIS:alg}. We employ this property to approximate the optimal annealing schedule in Section~\ref{algol:sec}.

To achieve accurate estimates with AIS, we have two problems to solve: the selection of the Markov transition operators $\trans$ and the selection of the intermediate distributions $\pseq$. As for transition operators, \citet{SohlDickstein:2012tk} recently proposed to implement $\trans$ with Hybrid Monte Carlo for enhanced mixing of Markov chains. 

As for intermediate distributions, there has been a long history of research \cite{Ogata:1989vja, Gelman:1998ux, Grosse:2013tf}. The design of intermediate distributions can be devided to two problems: the selection of an annealing path and the selection of an annealing schedule. An ``{\it annealing path}'' is a continuous parameterization of distributions $p_{\beta}$ with $\beta\in [0,1]$. Although AIS has its origin in statistical physics \cite{Iba:2001va}, $\beta$ need not be the inverse temperature, and any parameterization is possible. The most commonly used annealing path is the geometric path \cite{Neal:2001wea, Neal:1996uc, Salakhutdinov:2008eq, Tieleman:2008gw, Dauphin:2013um} although it is proved to be suboptimal in terms of the estimation accuracy \cite{Gelman:1998ux}. 
\citet{Gelman:1998ux} analyzed the relation between estimation errors and annealing paths, and derived the optimal annealing path that minimizes the errors. However, the optimal path suggested by \citet{Gelman:1998ux} is often intractable in practical applications. \citet{Grosse:2013tf} recently proposed the moment averaging path, which is still suboptimal but can be used in practical problems and results in better estimation accuracy than the geometric path.

Compared to annealing paths, little has been done on annealing schedules for AIS. An ``{\it annealing schedule}'' denotes a binning or quantization of $\beta\in [0,1]$. For the tempered transition method, which is deeply related to AIS, scheduling techniques for the geometric path are studied in terms of the acceptance rate \cite{Behrens:2012jo, Neal:1996uc}. However, these techniques are customized for the tempered transition and are not suitable for AIS. 
\citet{Grosse:2013tf} recently proposed a scheduling technique by formulating the problem as minimization of $\log\zB-\Exp\left[\log\wais\right]$. 
In this paper, we develop an alternative technique by variational minimization of a functional that dominates the variance of estimates i.e., $\var\left[\log\wais\right]$.

\section{Estimation Errors and An Annealing Schedule}
\label{analysis:sec}


For analyzing AIS, it is useful to assume a {\it perfect transition} condition where $\tran[\beta]$ returns independent samples of the previous ones \cite{Neal:2001wea, Grosse:2013tf}. This condition is an ideal situation where the mixing of Markov chains is very fast. 
Under this condition, $\log\wais$ can be regarded as a summation of $K$ independent random variables $(\log\pstrb[\beta_{k+1}](\vv) - \log\pstrb[\beta_k](\vv))$ where $\vv\sim\pk$. Therefore, as \citet{Neal:2001wea} suggests, $\log\wais$ approximately follows a normal distribution with lage $K$ as a consequence of the central limit theorem. The variance of $\log\wais$ can be computed as 
\begin{align}
\hspace{-4mm}
\var\left[\log\wais\right]&= \sum_{k=0}^K \varb[\beta_k]\left[\log\pstrb[\beta_{k+1}](\vv) - \log\pstrb[\beta_k](\vv) \right],
\label{aiserr:eq}
\end{align}
where $\varb$ denotes the variance w.r.t. $\pk[\beta]$.

\newcommand{\seq}{\mathrm{LINSPACE}}
\newcommand{\desol}{\mathrm{DESolve}}
\begin{algorithm}[t]
   \caption{VAROPT-AIS}
   \label{bais:alg}
\begin{algorithmic}
 \STATE {\bfseries Input:}  $\tilde{K}, \tilde{N}, K, N$
 \STATE Let $\hbetaseq$ be a $\tilde{K}$ uniformly spaced sequence of $[0,1]$
 \STATE Estimate $\{g(\tilde{\beta}_k)\}$ using $\AIS(\hbetaseq, \tilde{N})$.
 \STATE Compute $\betaseq = \desol(\{g(\tilde{\beta}_k)\})$.
 \STATE Estimate $\ze$ using $\AIS(\betaseq, N)$.
 \STATE {\bfseries Output:} $\ze$
\end{algorithmic}
\end{algorithm}

\begin{algorithm}[tb]
   \caption{Deceleration of schedules}
   \label{dec:alg}
\begin{algorithmic}
 \STATE {\bfseries Input:} schedule $\betaseq$, maximum delta $\dmax$, tolerance $\mathit{Tol}$
 \STATE Initialize $\dbeta = \beta_{k}-\beta_{k-1}$ for $k=1,\ldots,K$. 
   \REPEAT
   \STATE Initialize $noChange = true$.
   \FOR{$k=1$ {\bfseries to} $K-1$}
    \IF{$\dbeta > \dmax$} 
      \STATE $\dbeta \leftarrow \dmax$
    \ENDIF 
   \ENDFOR
   \STATE Compute $\mathit{Norm} = \sum_k\dbeta$
   \IF{$|\mathit{Norm} -1| < \mathit{Tol}$} 
     \STATE $noChange = false$
   \ENDIF
   \FOR{$k=1$ {\bfseries to} $K-1$}
     \STATE $\dbeta \leftarrow \dbeta/\mathit{Norm}$
   \ENDFOR
   \UNTIL{$noChange$ is $true$}
 \STATE $\beta_k = \sum_{\hat{k}=1}^{k}\dbeta[\hat{k}]$
 \STATE {\bfseries Output:} $\betaseq$
\end{algorithmic}
\end{algorithm}

\begin{figure}[t]
\vskip 0.2in
\begin{center}
\centerline{\includegraphics[width=0.8\columnwidth]{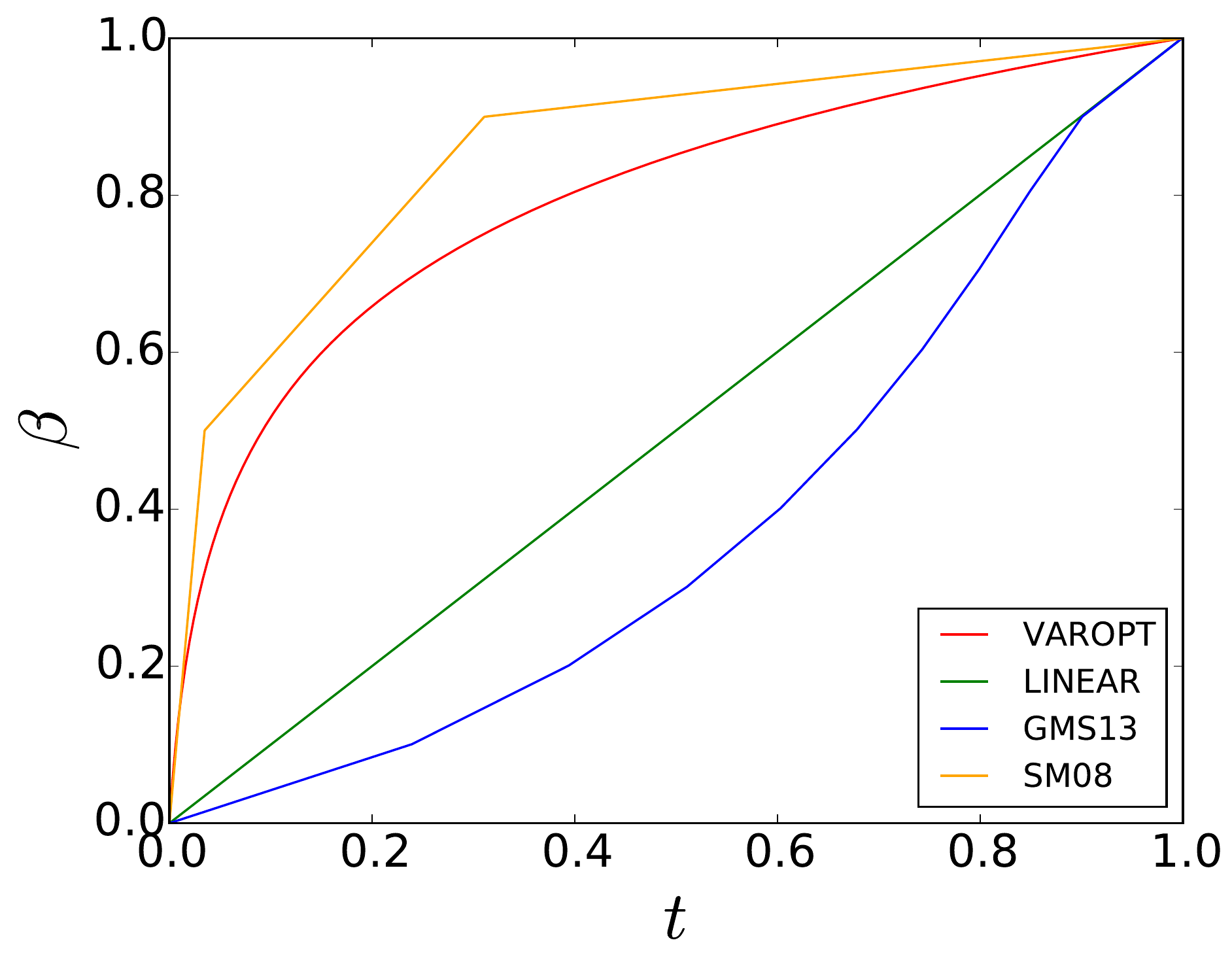}}
\caption{Comparison of annealing schedules for an RBM trained on MNIST by using PCD.}
\label{beta_t:fig}
\end{center}
\vskip -0.2in
\end{figure} 

\begin{figure*}[t]
\vskip 0.2in
\begin{center}
\hspace{1cm}PCD(20) \hspace{4cm} CD1(20)\hspace{4cm} CD25(20)\\
\includegraphics[width=0.63\columnwidth]{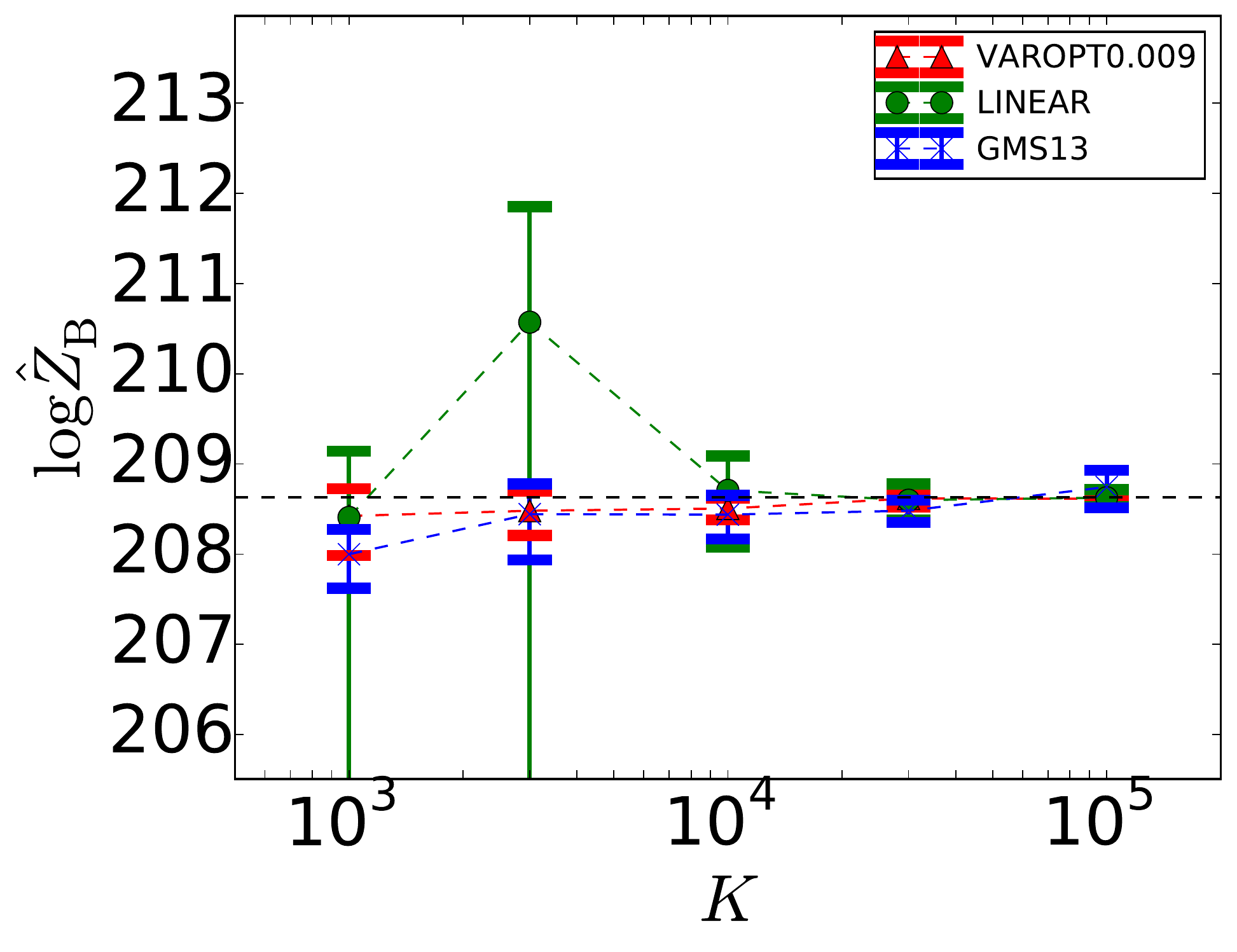}
\includegraphics[width=0.65\columnwidth]{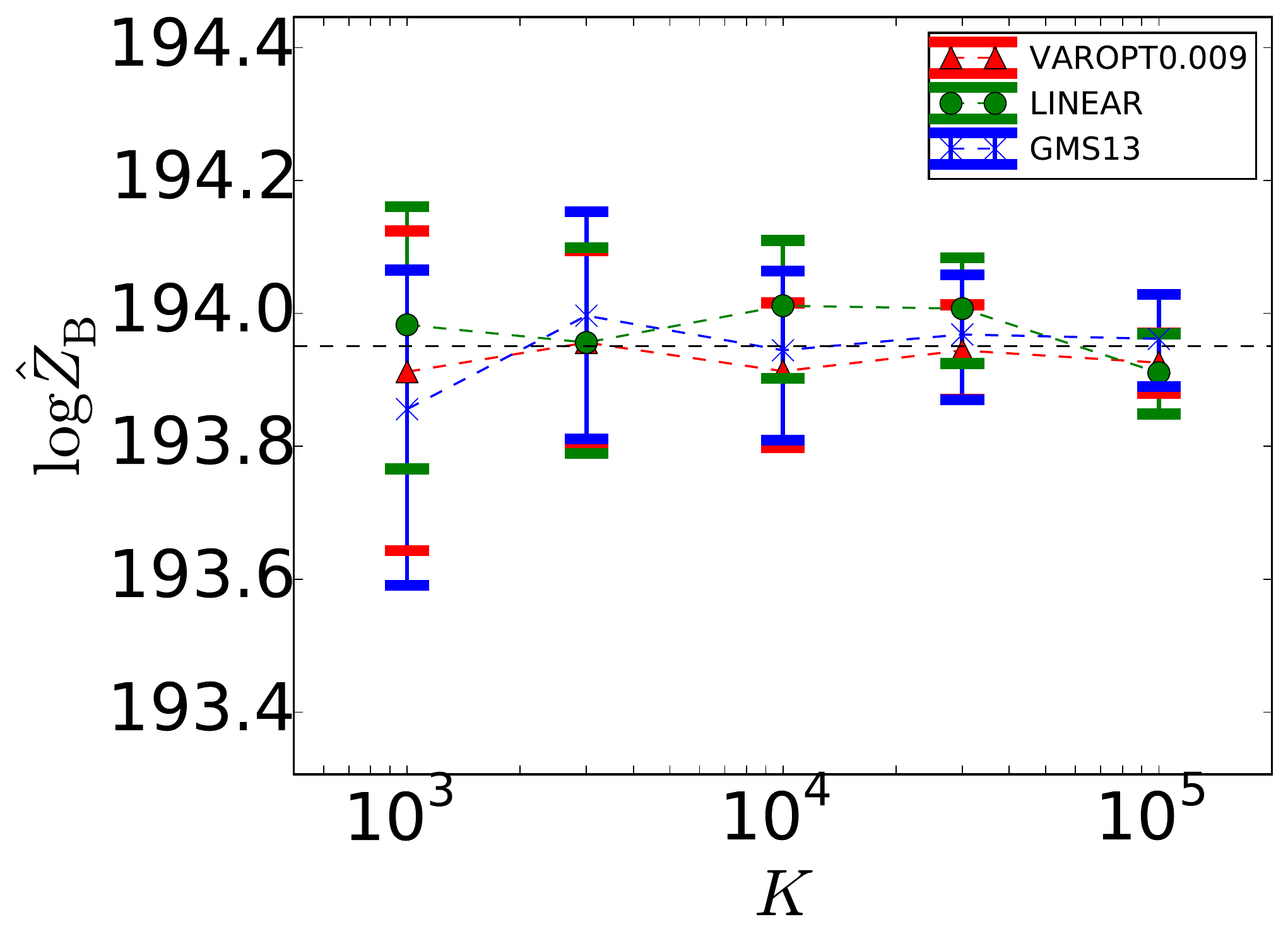}
\includegraphics[width=0.63\columnwidth]{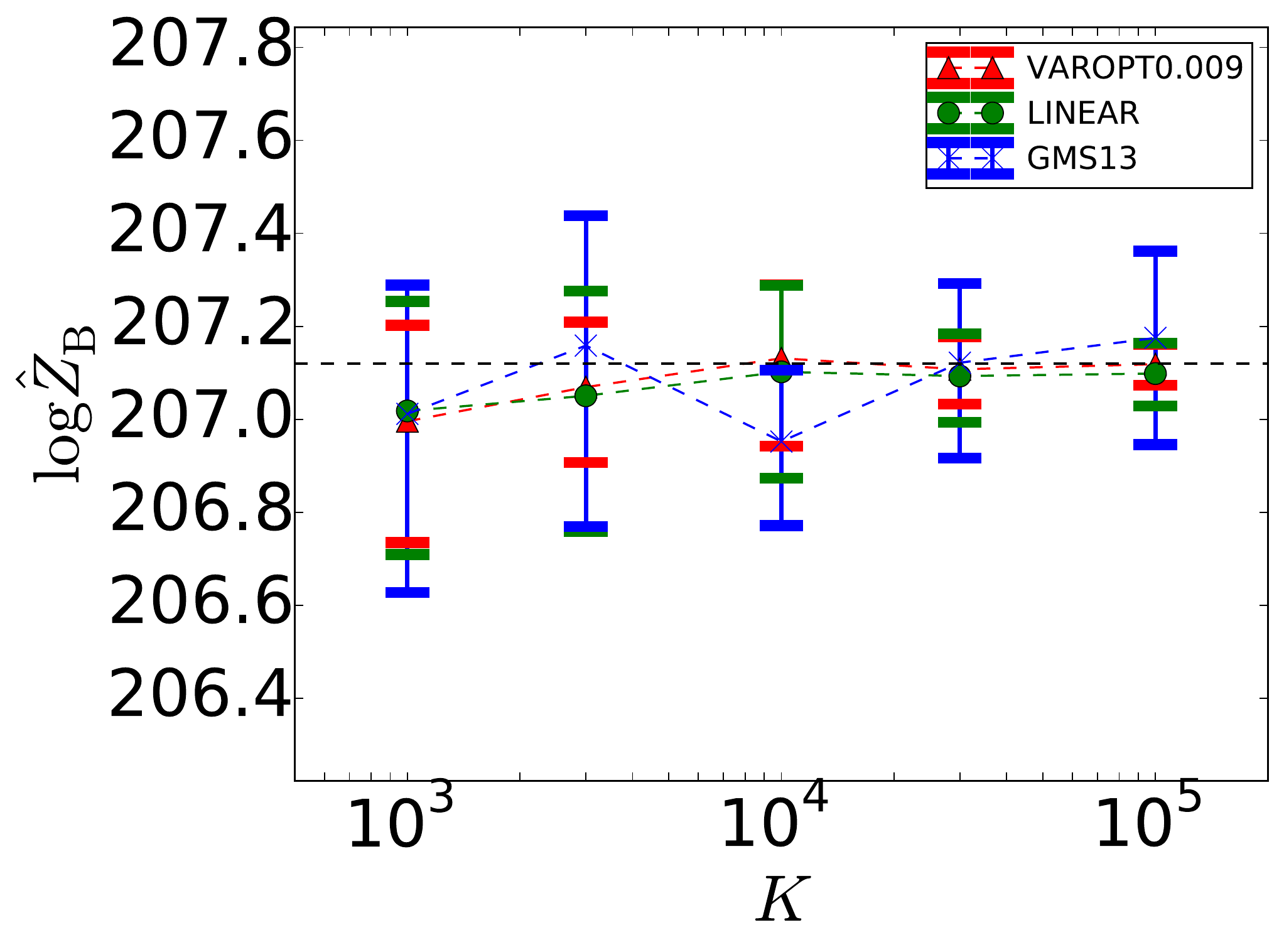}
\caption{$\log\ze$ for tractable RBMs as a function of $K$. Error bar shows $\pm3\sigma$ intervals of $\log\wais$. The black broken lines indicate the ground truth.}
\label{K_logZ_M20:fig}
\end{center}
\vskip -0.2in
\end{figure*} 
\begin{figure*}[t]
\vskip 0.2in
\begin{center}
\hspace{1cm}PCD(20) \hspace{4cm} CD1(20)\hspace{4cm} CD25(20)\\
\includegraphics[width=0.63\columnwidth]{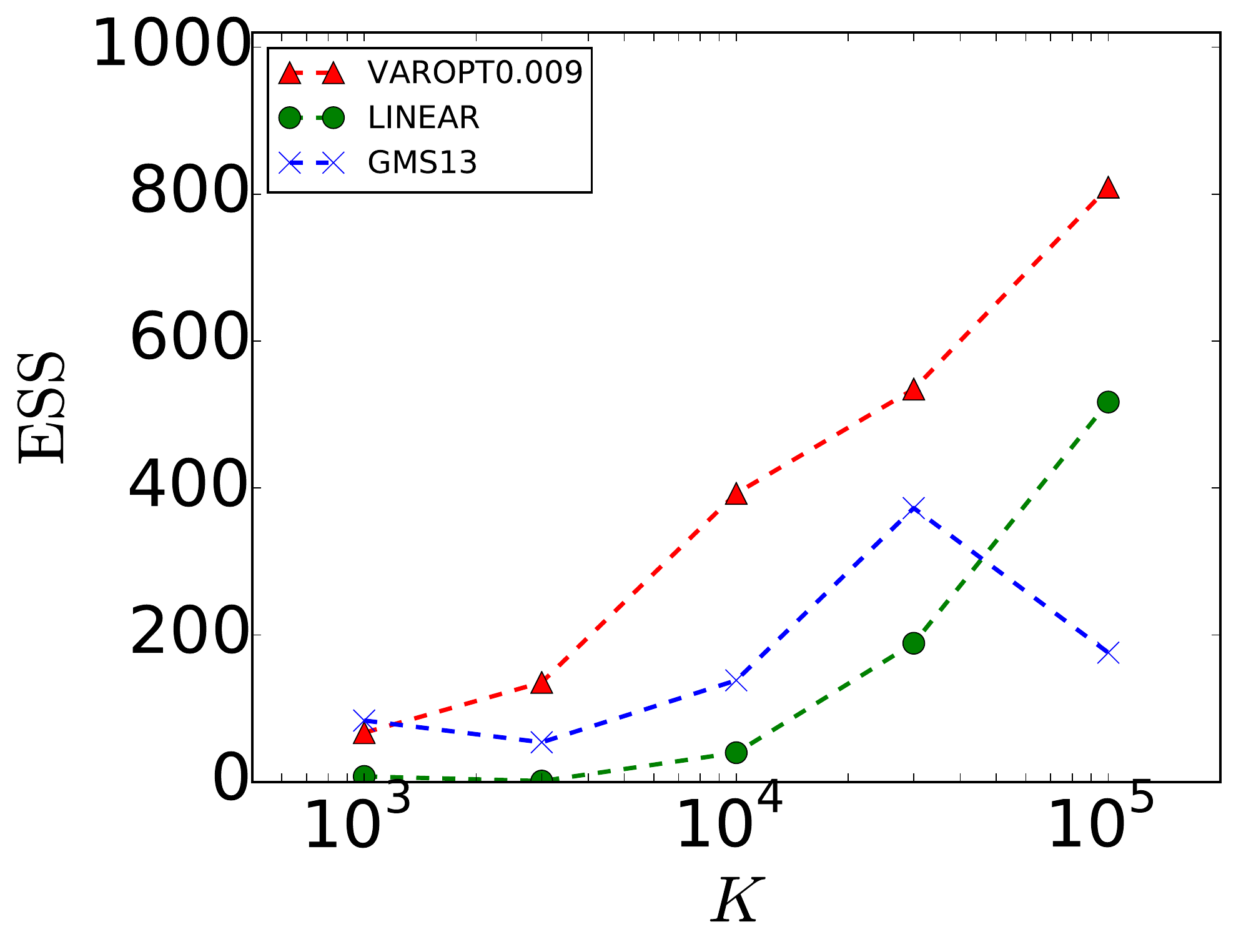}
\includegraphics[width=0.65\columnwidth]{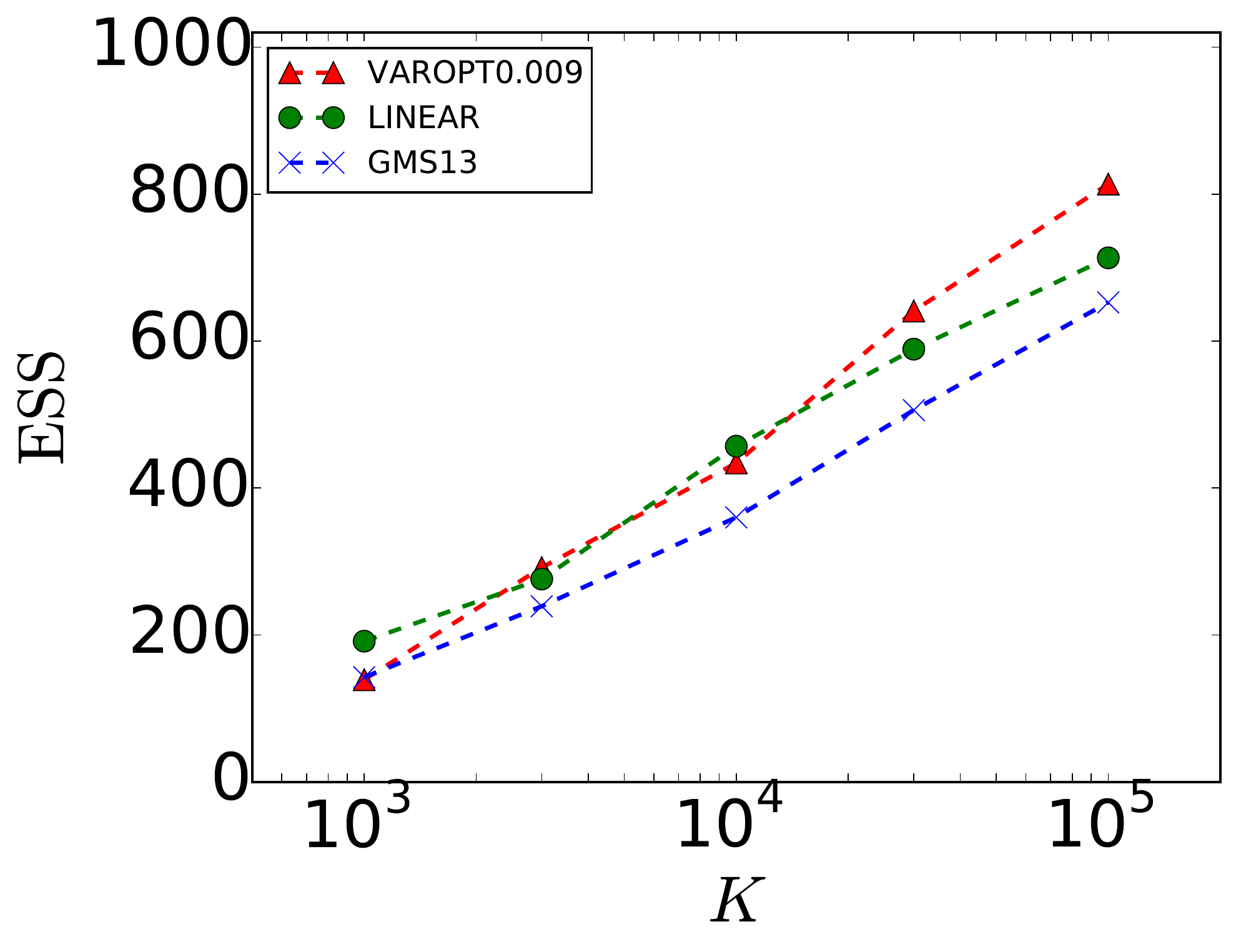}
\includegraphics[width=0.63\columnwidth]{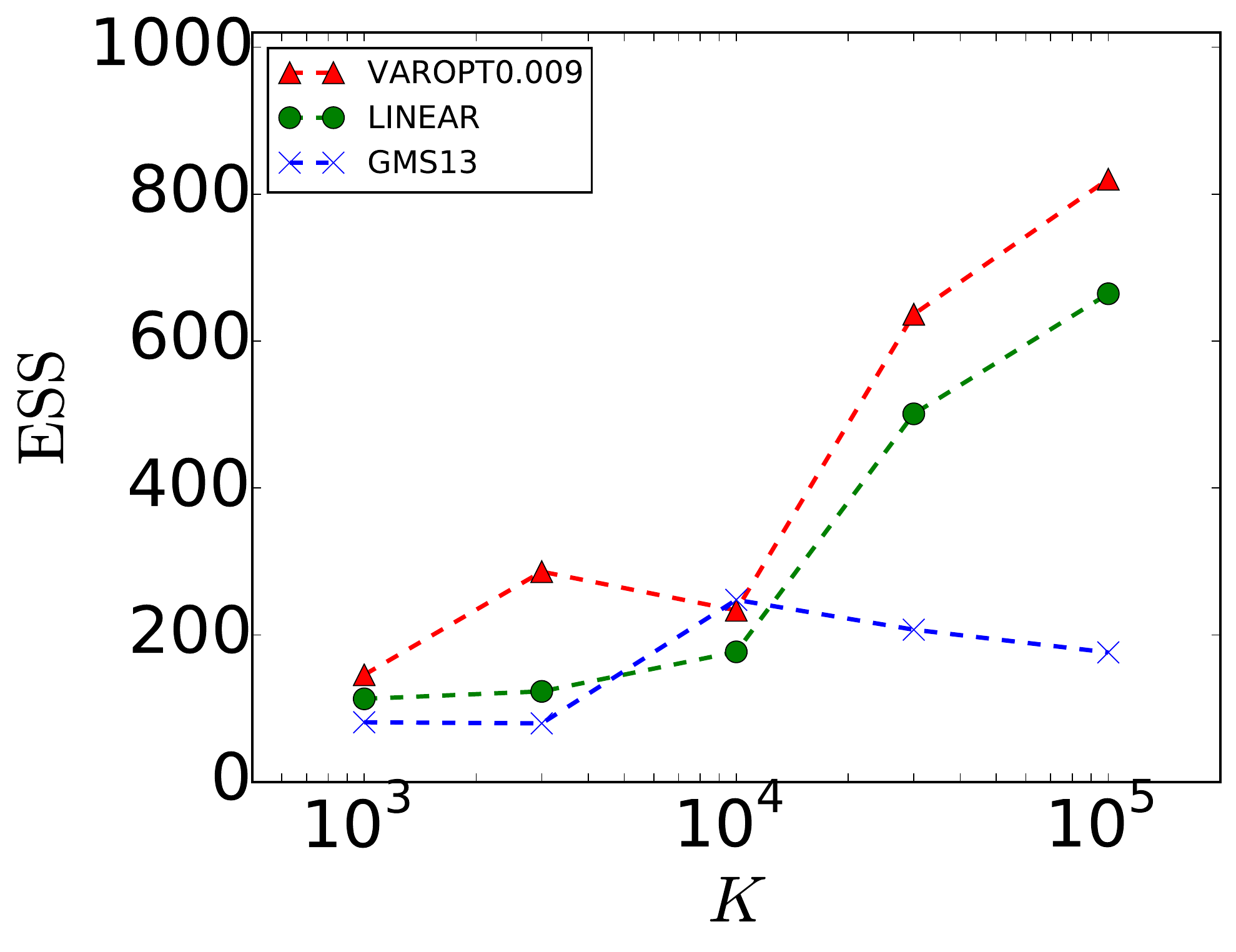}
\caption{Estimates of the ESS's plotted as a function of $K$.}
\label{K_ESS_M20:fig}
\end{center}
\vskip -0.2in
\end{figure*}

Because the variance becomes inversely proportional to $K$ as $K$ increases, we analyze the behavior of $K\var\left[\log\wais\right]$ for large $K$. By approximating the difference in the r.h.s. of \refeq{aiserr:eq} with a Taylor series up to fist order, we have the following approximation
\begin{align}
\hspace{-3mm}
K\var\left[\log\wais\right]
&\approx K\sum_{k=0}^K (\Delta\beta_k)^2 \varb[\beta_k]\left[\frac{\partial}{\partial \beta}\log\pstrb(\vv) \right].\label{approx:eq}
\end{align}
Assume that annealing schedule $\betaseq$ has a continuous limit i.e., $\beta_k = \beta(k/K)$ with some smooth function $\beta(t)$ defined on $t\in[0,1]$. Because the error caused by the approximation vanishes under this assumption as $K\rightarrow\infty$, the scaled variance asymptotically approaches a functional $\obj$. 


\begin{theorem}
Assume perfect transitions. 
Assume that $\betaseq$ are composed as $\beta_k = \beta(\tk)$ where $\beta(t)$ is a smooth function ($\beta(t)\in{\cal C}^2$) defined on $t\in[0,1]$ and $\tk=k/K$. 
Then as $K\rightarrow\infty$ the AIS estimation error behaves as:
\begin{align}
K\var\left[\log\wais\right]\rightarrow \obj \triangleq \int_0^1\vbeta^2 g(\beta) \dd t, 
\end{align}
where $\vbeta$ denotes the derivative of $\beta(t)$, i.e., $\frac{\dd\beta(t)}{\dd t}$, and $g(\beta)$ is a function defined as $g(\beta)\triangleq\varb\left[\frac{\partial}{\partial \beta}\log\pstrb(\vv)\right]$. [See supplementary material for proof.] \label{limit:th} 
\end{theorem}

\begin{table*}[t]
\begin{center}
\caption{Estimates of the partition functions and the ESS's for tractable RBMs. The ground truth of the estimate $\log\zB$ is also reported. All the figures are obtained with $K=100,000$}
\label{M20:tab}
 \begin{tabular}{lccrccrccr}
 \hline 
&  \multicolumn{3}{c}{PCD(20)}&  \multicolumn{3}{c}{CD1(20)}&  \multicolumn{3}{c}{CD25(20)}\\
\cmidrule(r){2-4}\cmidrule(r){5-7}\cmidrule(r){8-10}
schedule     	&$\log\zB$	&$\log\ze$	&ESS	&$\log\zB$	&$\log\ze$	&ESS	&$\log\zB$	&$\log\ze$	&ESS \\ \hline
         VAROPT& 208.63	&208.629	&  783	&193.951	&194.117	&   87	&207.12	&207.136	&  776 \\ 
    VAROPT0.009& 	&208.616	&  {\bf 809}	&	&193.926	& {\bf 814}&	&207.119	& {\bf 820} \\ 
    VAROPT0.006& 	&208.643	&  668	&	&193.937	&  803	&	&207.148	&  751 \\ 
    VAROPT0.003& 	&208.626	&  749	&	&193.956	&  797	&	&207.136	&  617 \\ \hline 
         LINEAR& 	&208.626	&  517	&	&193.911	&  713	&	&207.099	&  664 \\ 
          GMS13& 	&208.745	&  176	&	&193.962	&  653	&	&207.175	&  176 \\ \hline 
\end{tabular}
\end{center}
\end{table*}

\begin{figure*}[ht]
\vskip 0.2in
\begin{center}
{\large$K=1,000$ \hspace{3.5cm} $10,000$\hspace{4cm} $100,000$}\\
\includegraphics[width=0.63\columnwidth]{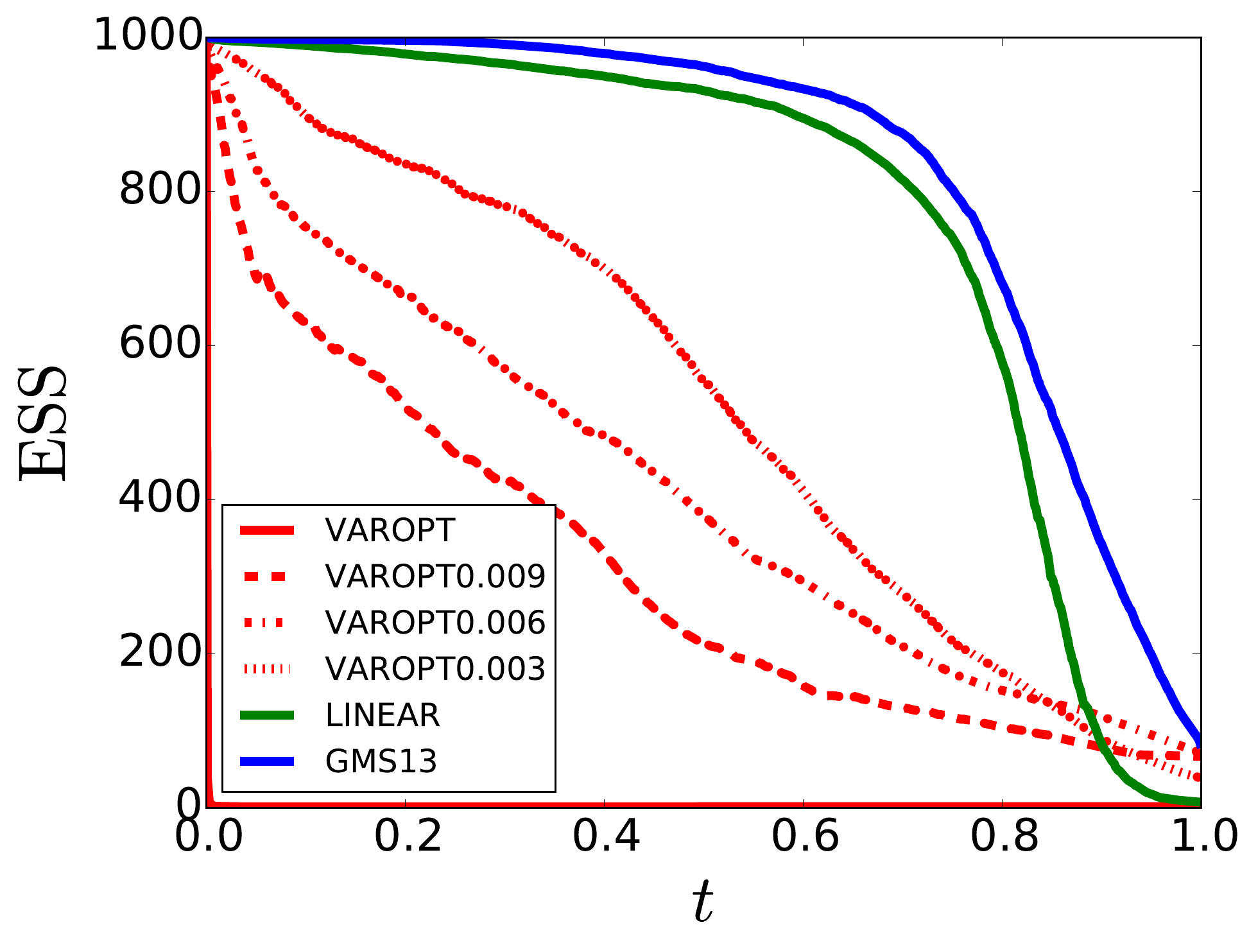}
\includegraphics[width=0.65\columnwidth]{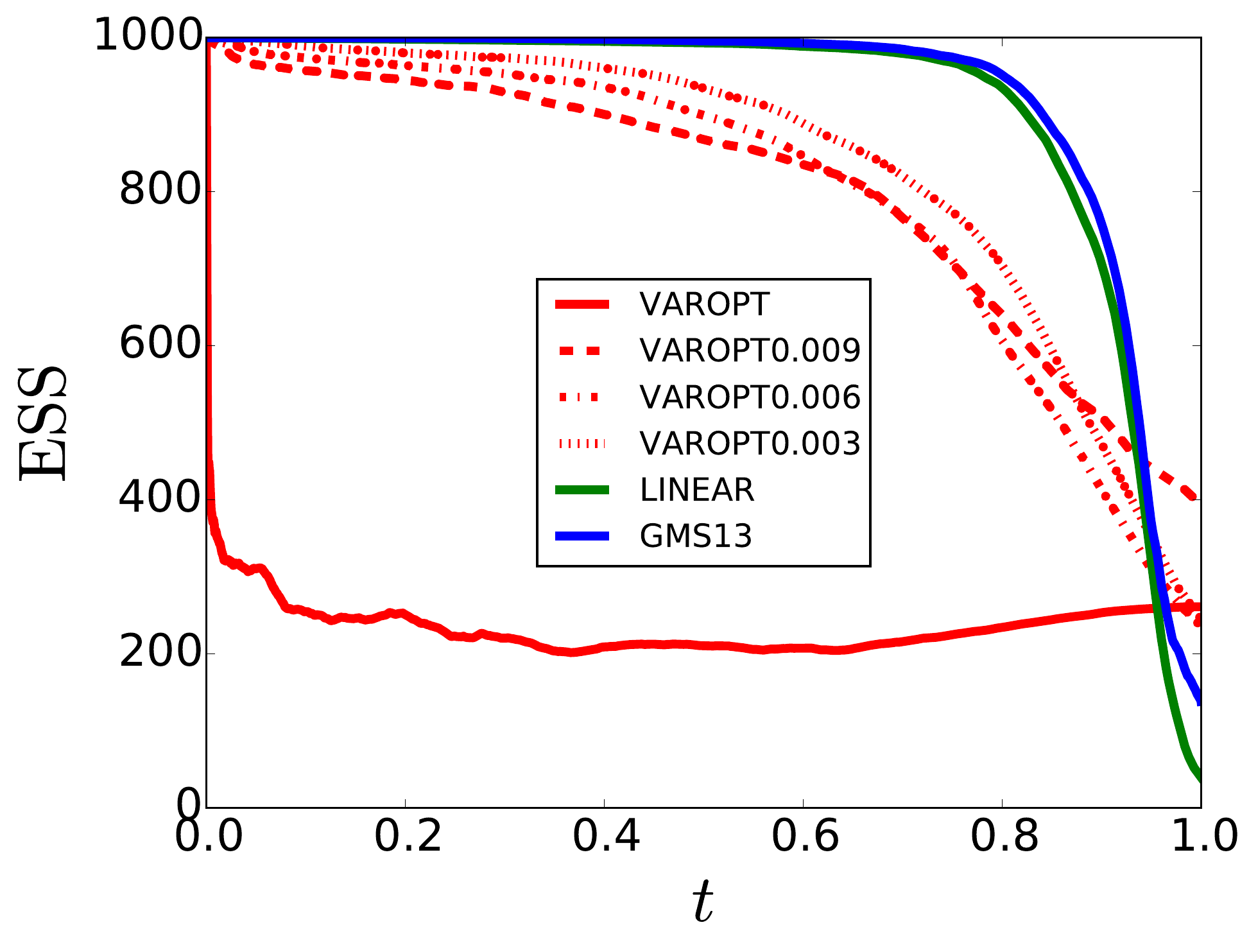}
\includegraphics[width=0.63\columnwidth]{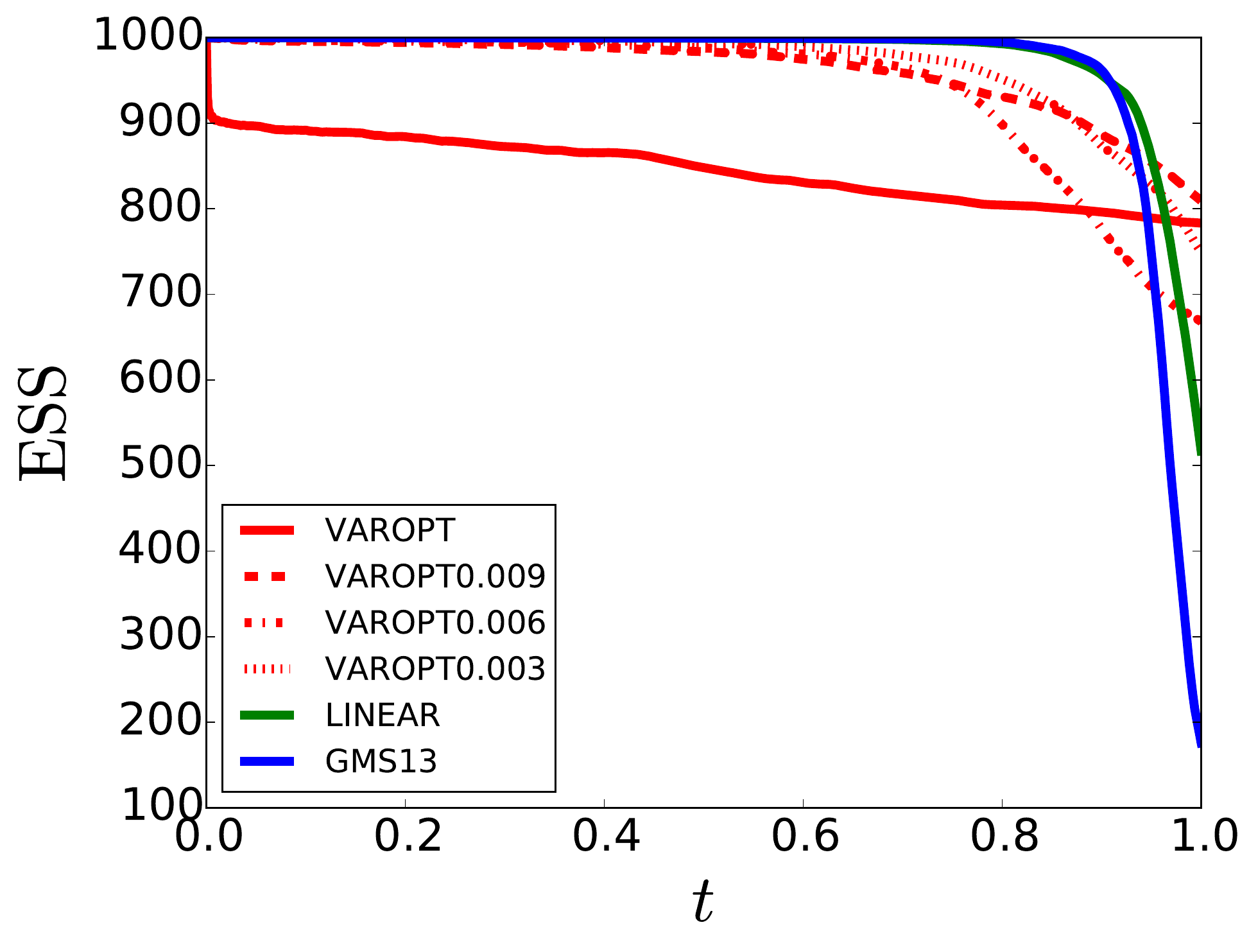}
\caption{Evolution of the ESS's on various $K$. These plots are computed with PCD(20)}
\label{k_ESS_M20:fig}
\end{center}
\vskip -0.2in
\end{figure*} 

Here the problem of finding the optimal schedule that minimizes the estimation error is formulated as a variational minimization problem of the functional $\obj$ w.r.t. $\beta(\cdot)$. 
From Euler-Lagrange equation \cite{Bishop:2006ui}, we derived the following differential equation that the optimal schedule obeys:
\begin{align}
 \abeta + \frac{\vbeta^2}{2}\frac{\dd }{\dd \beta}\log g(\beta)  = 0.\label{de:eq}
\end{align}



\section{Numerical Search for The Optimal Schedule}
\label{algol:sec}
By numerically solving \refeq{de:eq} with boundary conditions $\betat[0]=0$ and $\betat[1]=1$, we can find the optimal schedule. A problem here is that $g(\beta)$ is intractable. To overcome this difficulty, we propose VAROPT-AIS algorithm listed in \refalg{bais:alg} where we perform AIS twice; $g(\beta)$ is estimated by the first (cheap) execution of AIS with the linear scheduling, and $\log\zB$ is estimated by the second (expensive) execution of AIS with a schedule computed from the first execution. A key idea of VAROPT-AIS is {\it to execute cheap AIS first to roughly survey the terrain through the annealing path i.e., $g(\beta)$, and then execute expensive AIS to gain thorough estimation}.

In VAROPT-AIS, we use the method of fixed point iteration \cite{Kelley:1995ux} to solve the differential equation of \refeq{de:eq} (labeled as DESolve in \refalg{bais:alg}). Because the l.h.s. of \refeq{de:eq} does not directly depend on $g(\beta)$, we perform numerical differentiation to approximate $\frac{\dd }{\dd \beta}\log g(\beta)$ as preprocessing. We also perform convolutional smoothing of $g(\beta)$ estimates to remove less important noises. 

Because \refeq{de:eq} is derived based on the assumption of perfect transitions, the solution of \refeq{de:eq} can have large $\dbeta =\beta_{k}-\beta_{k-1}$ that can impede the mixing of Markov chains. This can damage the estimation accuracy of AIS. To ease this effect, we optionally decelerate an annealing schedule s.t. $\max\dbeta \le \dmax$ with \refalg{dec:alg}. This heuristic algorithm sequentially clips $\dbeta$ by $\dmax$ and stretches all $\dbeta$ to compensate the error caused by clipping.

\section{Remarks}
The methodology developed in this paper can be applied to various kinds of stochastic models to which AIS is applicable. Nevertheless, we are mainly interested in RBMs and only perform experiments on RBMs in this paper. 

Also note that our method can be combined with various kinds of established techniques for AIS. First, the proposed method can be combined with HAIS \cite{SohlDickstein:2012tk} because the selection of an annealing schedule is independent of the implementation of Markov transitions. Second, the proposed method can be used to schedule various types of annealing paths possibly including the moment averaging path with geometric interpolation \cite{Grosse:2013tf}. Finally, the proposed method will easily be combined with a technique for tuning a proposal distribution \cite{Kiwaki:n5j3qBs2}. 

\refFig{beta_t:fig} compares an annealing schedule by our method (labeled as {\bf VAROPT}) with those by several others: ({\bf LINEAR}) the linear schedule that corresponds to $\beta(t)=t$; ({\bf GMS13}) a scheme suggested by \citet{Grosse:2013tf}; and ({\bf SM08}) a heuristic schedule suggested by \citet{Salakhutdinov:2008eq}. Several interesting points can be seen from these plots. First, GMS13 is largely different from VAROPT and is rather similar to the linear schedule. This clearly shows that our objective $\obj$ is intrinsically different from the objective proposed by \citet{Grosse:2013tf}. Second, VAROPT is similar to the heuristic schedule by \citet{Salakhutdinov:2008eq}. This remarkable coincidence suggests that our proposal possibly automates expensive heuristic search of annealing schedules with human hands.

\begin{figure*}[ht]
\vskip 0.2in
\begin{center}
\hspace{1cm}PCD(500) \hspace{4cm} CD1(500)\hspace{4cm} CD25(500)\\
\includegraphics[width=0.63\columnwidth]{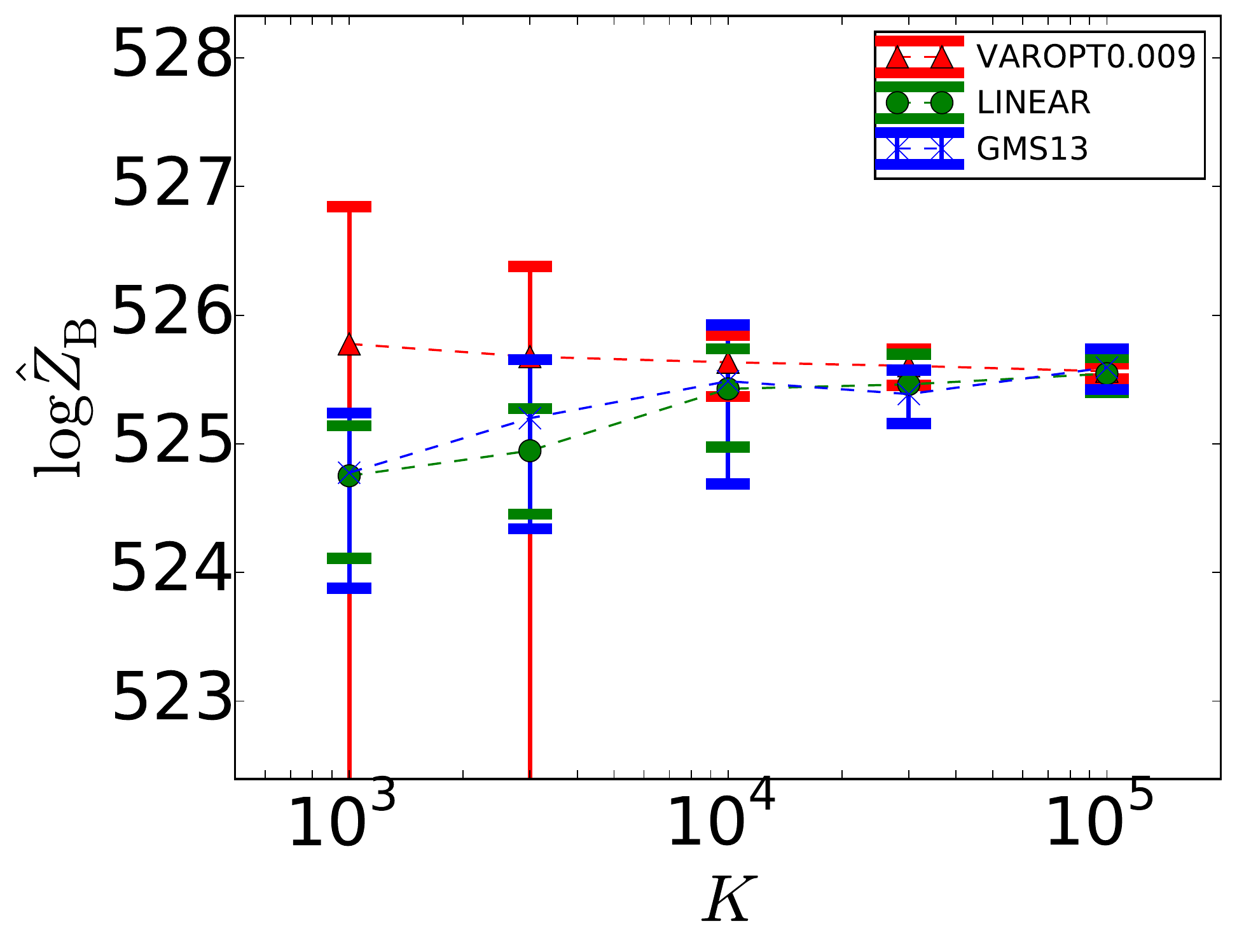}
\includegraphics[width=0.65\columnwidth]{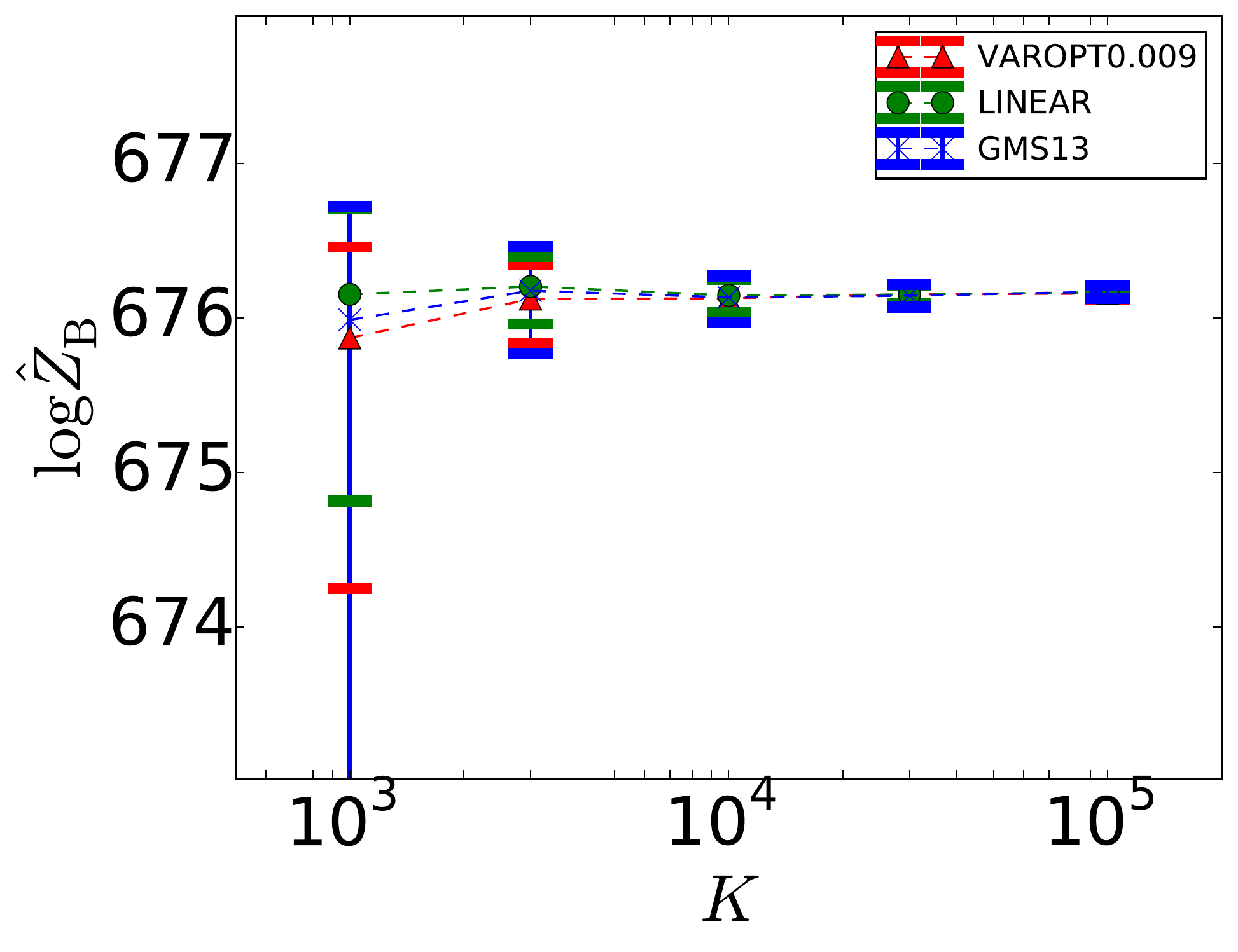}
\includegraphics[width=0.63\columnwidth]{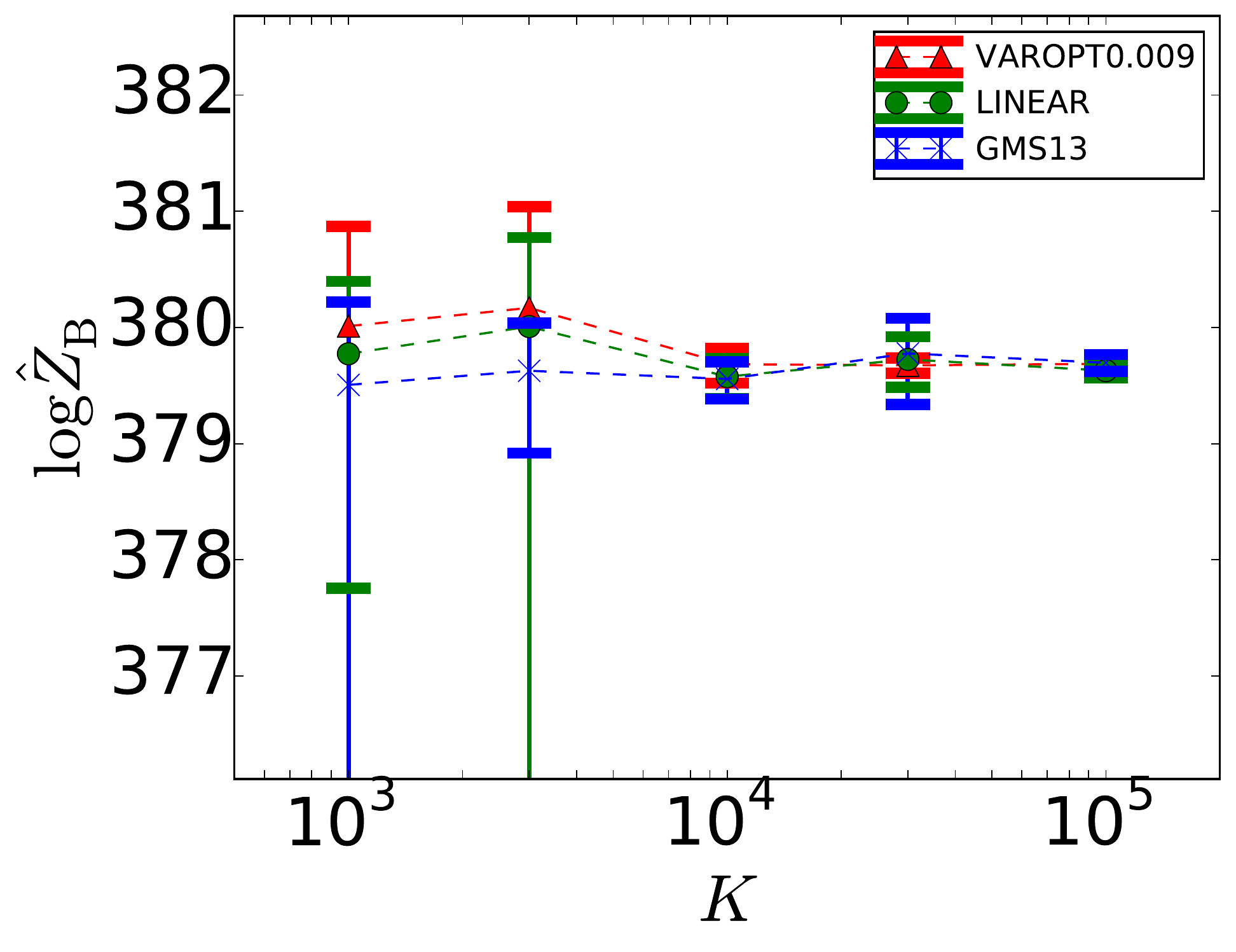}
\caption{$\log\ze$ for intractable RBMs as a function of $K$. Error bar shows $\pm3\sigma$ intervals of $\log\wais$. }
\label{K_logZ_M500:fig}
\end{center}
\vskip -0.2in
\end{figure*} 

\begin{figure*}[ht]
\vskip 0.2in
\begin{center}
\hspace{1cm}PCD(500) \hspace{4cm} CD1(500)\hspace{4cm} CD25(500)\\
\includegraphics[width=0.63\columnwidth]{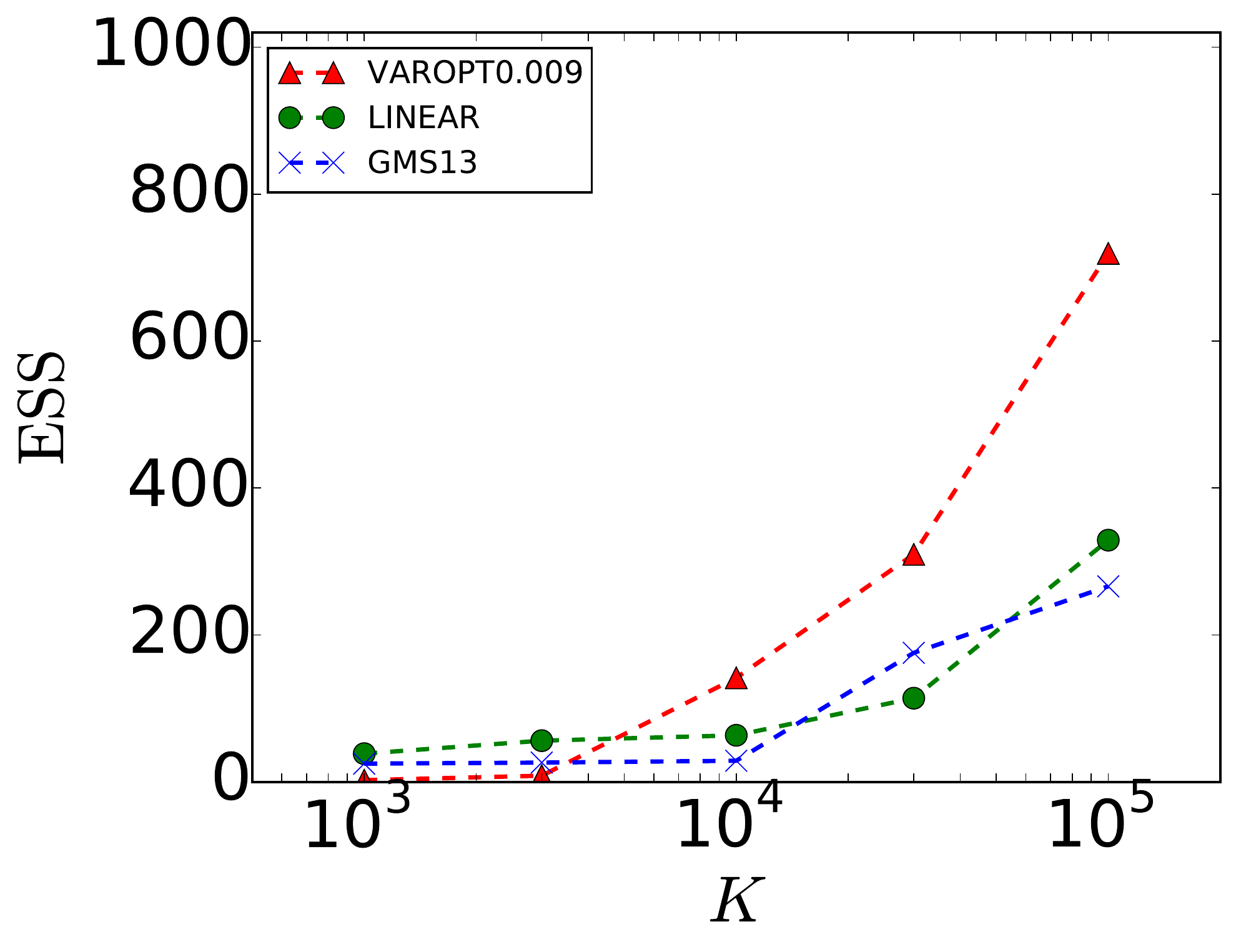}
\includegraphics[width=0.65\columnwidth]{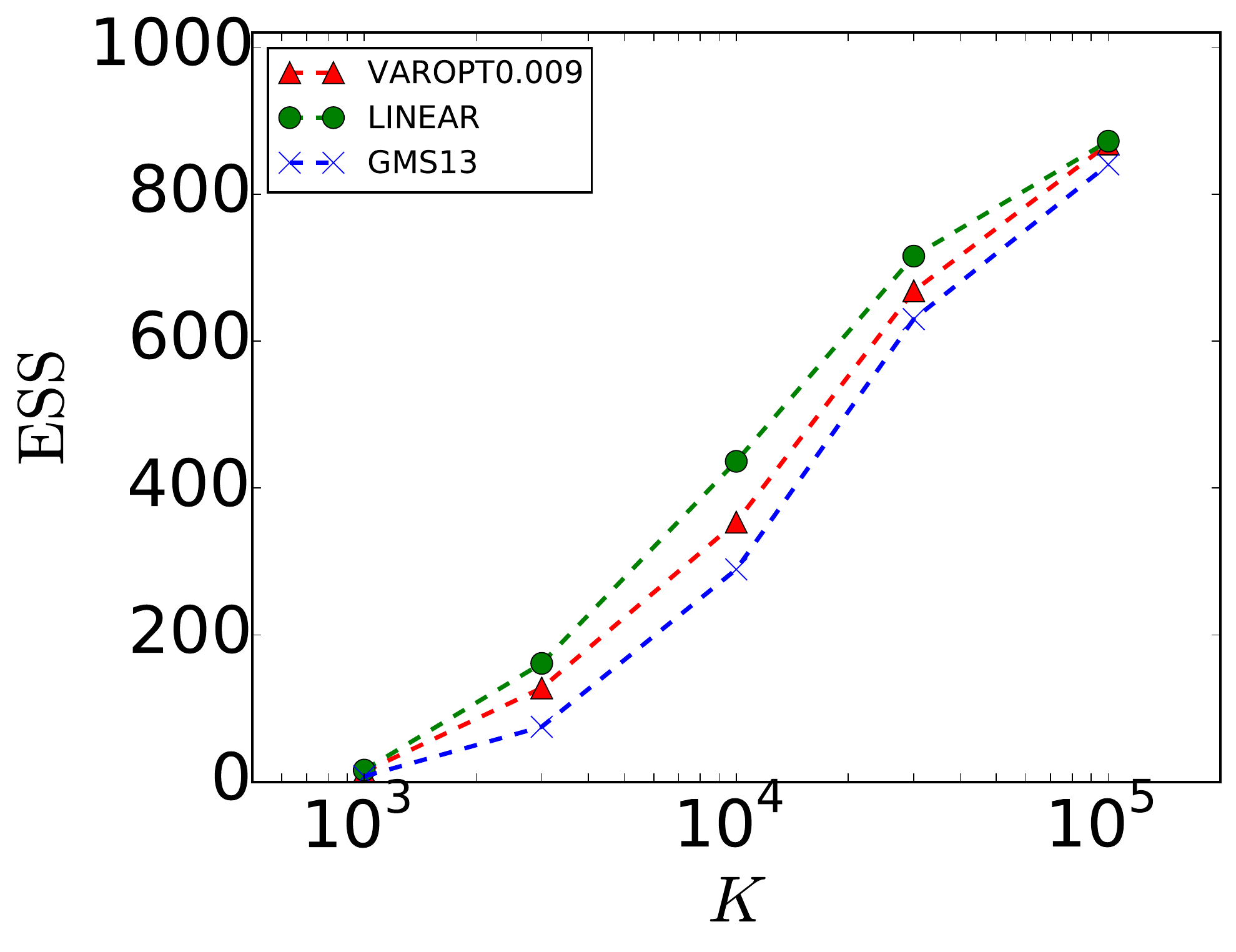}
\includegraphics[width=0.63\columnwidth]{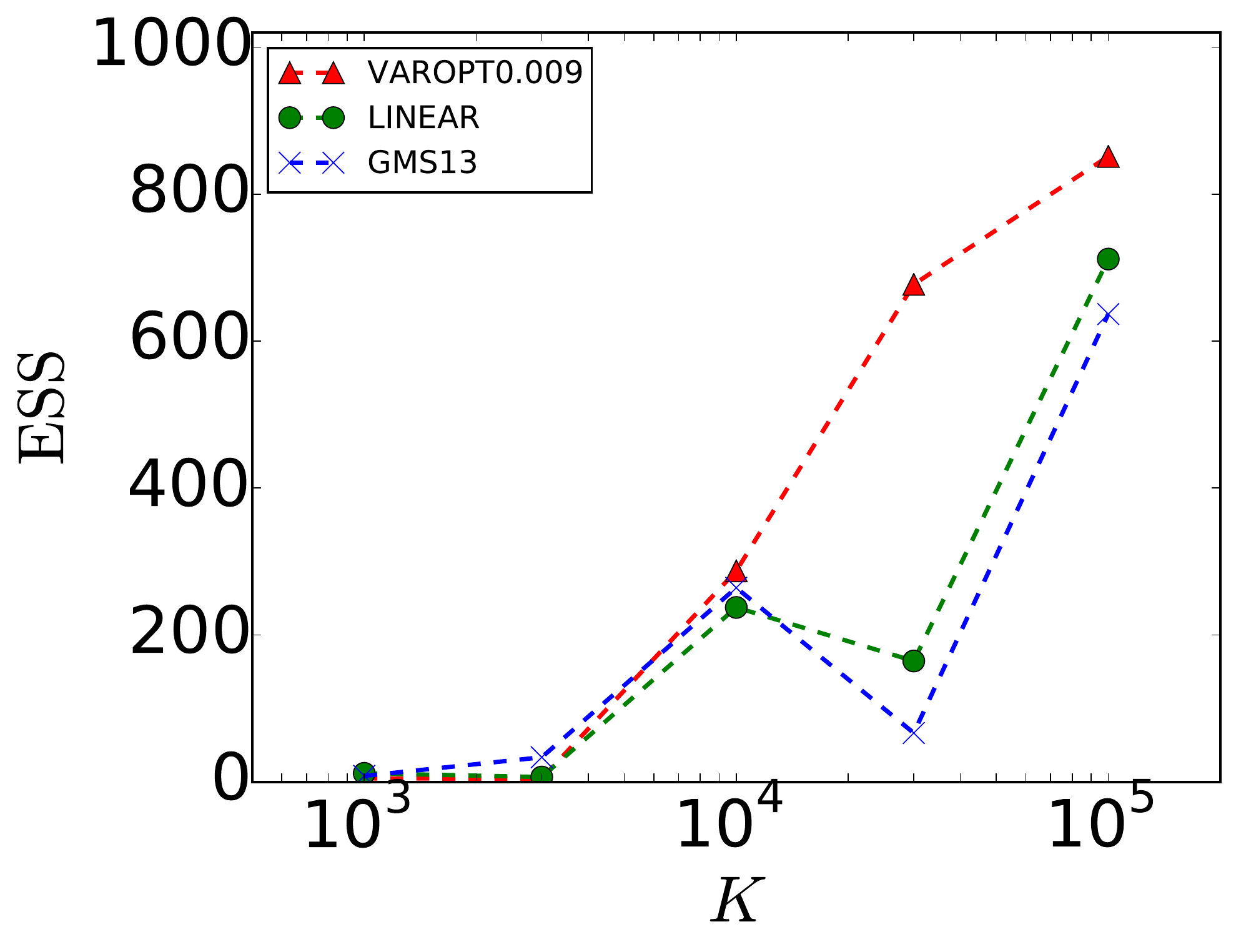}
\caption{ESS estimates for intractable RBMs as a function of $K$.}
\label{K_ESS_M500:fig}
\end{center}
\vskip -0.2in
\end{figure*}

\section{Experiments}

To demonstrate the benefits of the proposed method, we performed partition function estimation for several RBMs with various annealing schedules. We evaluated scheduling schemes with respect to two measures: ESS estimates and $\log\ze$. As \citet{Neal:2001wea} warns, ESS estimates can be misleading if AIS fails to find important major modes of $\pB$, and therefore one should be careful when reporting the estimates. In our experiments, however, we regard that ESS estimates are reliable because the partition function estimates seem reliable in most cases from comparison with the ground truth or from comparison with estimates by different schemes. 

RBMs were trained on MNIST by using three training algorithms: ({\bf PCD}) persistent contrastive divergence \cite{Tieleman:2008gw}, ({\bf CD1}) contrastive divergence (CD) with 1 step of state update, and ({\bf CD25}) CD with 25 steps \cite{Hinton:2002wb}. We label RBMs with the training algorithm and the number of hidden units; for example, PCD(500) denotes an RBM with 500 hidden units trained by PCD.

All the executions of AIS followed the geometric path. We fixed the number of AIS runs as $N=1,000$ and explored various magnitudes of $K\in 10^{[3,5]}$.

We mainly compared following three scheduling techniques: GMS13, LINEAR, and VAROPT. VAROPT schedules were computed with $\tilde{N}=100$ and $\tilde{K}=1,000$. Note that the computation required to gain VAROPT schedules was not heavy and negligible compared to the cost of the following main execution of AIS. 
In addition to simple VAROPT, we also tested decelerated VAROPT schedules with $\dmax \in \{0.003, 0.006, 0.009\}$. Decelerated schedules are labels as VAROPT$\dmax$. 

GMS13 schedules are determined using 10 different points of RBM parameters (knots) on the geometric path. On each knot, we estimated the moments of the RBM using 1,000 Markrov chains with 6,000 updates including 1,000 steps of burn-in updates. 

\subsection{Experiments with Small Tractable RBMs}
We first show results with RBMs that have only 20 hidden units. Note that we can compute the exact value of $\log\zB$ for these RBMs by summing all the $2^{20}$ possible states of hidden units. For training RBMs, we used a fixed learning rate and performed 250,000 parameter updates.

The results of estimates are summarized in \reftab{M20:tab}. It is remarkable to note that VAROPT0.009 achieves the highest ESS's for all the RBMs. Estimates of $\log\zB$ and the ESS's are plotted as a function of $K$ in Figs.~\ref{K_logZ_M20:fig} and \ref{K_ESS_M20:fig}. Note that the proposed method is solely represented by VAROPT0.009 in these plots. From these plots, it can be observed that VAROPT0.009 achieves the smallest estimation errors and the greatest ESS's in most of the cases, especially with large $K$.

To better understand the behavior of VAROPT, we computed ESS estimates with on-the-fly AIS weights as shown in \reffig{k_ESS_M20:fig}. 
Note that such on-the-fly ESS's are valid statistics because on-the-fly AIS weights can be used to estimate the statistics of the intermediate distributions. Because the estimation error is accumulated throughout annealing (as \refeq{aiserr:eq} suggests), monitoring on-the-fly ESS estimates helps us to understand the characteristics of annealing schedules. It can be seen from \reffig{k_ESS_M20:fig} that VAROPT has a steep drop in the ESS's at very the beginning of the annealing. Is is also shown that deceleration effectively relaxes this problem to yield higher ESS's. 
We understand that this sudden drop in the ESS's is due to poor mixing of Markov chains because the drop becomes smaller with larger value of $K$. Therefore, the larger $K$ becomes, the better estimation accuracy VAROPT enjoys. 


\begin{table*}
\begin{center}
\caption{Estimates of the partition functions and the ESS's for intractable RBMs. All the figures are obtained with $K=100,000$}
\label{M500:tab}
 \begin{tabular}{lcrcrcr}
 \hline 
&  \multicolumn{2}{c}{PCD(500)}&  \multicolumn{2}{c}{CD1(500)}&  \multicolumn{2}{c}{CD25(500)}\\
\cmidrule(r){2-3}\cmidrule(r){4-5}\cmidrule(r){6-7}

schedule     	&$\log\ze$	&ESS	&$\log\ze$	&ESS	&$\log\ze$	&ESS \\ \hline
         VAROPT& 525.55	&  726	&676.149	&  865	&379.68	&  545 \\ 
    VAROPT0.009& 525.564	&  719	&676.158	&  868	&379.687	&  851 \\ 
    VAROPT0.006& 525.529	& {\bf 728} &676.169	&  855	&379.682	& {\bf 852} \\ 
    VAROPT0.003& 525.548	&  661	&676.13	& {\bf 873} &379.68	&  778 \\ \hline 
         LINEAR& 525.545	&  329	&676.167	&  872	&379.628	&  712 \\ 
          GMS13& 525.593	&  266	&676.17	&  840	&379.696	&  637 \\ \hline 

 \end{tabular}
\end{center}
\end{table*}

\subsection{Experiments with Intractable RBMs}

We next report estimation on intractable RBMs with 500 hidden units. RBMs were trained using randomly sampled hyperparameters such as the number of training epochs, learning rates, and L2 regularization. 

Estimates of $\log\zB$ and the ESS's are plotted in Figs.~\ref{K_logZ_M500:fig} and \ref{K_ESS_M500:fig}. 
\reftab{M500:tab} shows the $\log\zB$ and ESS estimates for $K=100,000$. The scores by (decelerated) VAROPT here look less appealing than for tractable RBMs. Especially, (decelerated) VAROPT exhibits large estimation errors for small $K$. This is possibly due to poorer mixing of RBMs with a larger number of hidden units. Nevertheless, estimation errors with (decelerated) VAROPT are rapidly reduced as $K$ increases. Thus, decelerated VAROPT schedules achieve greater ESS's than the conventional scheduling schemes for all the RBMs with $K=100,000$ as in \reftab{M500:tab}.

\section{Conclusion}
We pursued a problem of determining the optimal annealing schedule for AIS. Assuming perfect transition, we derived a functional that dominates the estimation error and formulated the problem as a variational minimization problem. We developed a numerical scheme to solve this variational problem and implemented a practical algorithm to approximate the optimal annealing schedule. We performed experiments and demonstrated that the proposed algorithm achieved better estimation accuracy than conventional schemes in most cases with a large number of intermediate distributions. 

 \section*{Acknowledgments} 
 This research is supported by JSPS Grant-in-Aid for JSPS Fellows (145500000159).  We thank Tomoya Takeuchi for valuable discussion. 

\bibliography{example_paper}
\bibliographystyle{icml2015}

\appendix
\setcounter{theorem}{0}
\section{Derivation of \refeq{approx:eq}}
By Taylor series expansion of $\log\pstrb(\vv)$ w.r.t. $\beta$, the variance $\var\left[\log\wais\right]$ can be written  as
\begin{align}
 &\var\left[\log\wais\right] =\sum_{k=0}^K \varb[\beta_k]\left[\log\pstrb[\beta_{k+1}](\vv) - \log\pstrb[\beta_k](\vv) \right]\label{ten:eq}\\
& = \sum_{k=0}^K \varb[\beta_k]\left[\frac{\partial}{\partial \beta}\log\pstrb(\vv)\dbeta + \delta(\vv, \beta_k)O(\dbeta^2) \right],\label{elev:eq}
\end{align}
where we defined $\dbeta=\beta_{k+1} - \beta_{k}$, and the coefficients for the higher order terms are represented by $\delta(\vv, \beta_k)$. Because $\dbeta$ does not depend on $\vv$, $K\var\left[\log\wais\right]$ can be further rewritten as 
\begin{align}
 &\var\left[\log\wais\right] \cnteq
&= \sum_{k=0}^K \left\{\dbeta^2 \varb[\beta_k]\left[\frac{\partial}{\partial \beta}\log\pstrb(\vv) \right] +\delta(\beta_k)O(\dbeta^3)\right\}\label{approx_detail:eq}
\end{align}
where $\delta(\beta_k) = 2\mathrm{Cov}_{\beta_k}\left[\delta(\vv, \beta_k), \frac{\partial}{\partial \beta}\log\pstrb(\vv)\right]$ with $\mathrm{Cov}_{\beta_k}$ being the covariance operator w.r.t. $\pb$. Neglect of the second term of the r.h.s. yields \refeq{approx:eq}.

\section{Proof of \refth{limit:th}}

\begin{theorem}
Assume perfect transitions. 
Assume that $\betaseq$ are composed as $\beta_k = \beta(\tk)$ where $\beta(t)$ is a smooth function ($\beta(t)\in{\cal C}^2$) defined on $t\in[0,1]$ and $\tk=k/K$. 
Then as $K\rightarrow\infty$ the AIS estimation error behaves as:
\begin{align}
K\var\left[\log\wais\right]\rightarrow \obj \triangleq \int_0^1\vbeta^2 g(\beta) \dd t, 
\end{align}
where $\vbeta$ denotes the derivative of $\beta(t)$, i.e., $\frac{\dd\beta(t)}{\dd t}$, and $g(\beta)$ is a function defined as $g(\beta)\triangleq\varb\left[\frac{\partial}{\partial \beta}\log\pstrb(\vv)\right]$. 
\end{theorem} 

\begin{proof}
 From \refeq{approx_detail:eq}, the scaled variance is written as
\begin{align}
 K\var\left[\log\wais\right] = &\frac{1}{K}\sum_{k=0}^K \left(K\dbeta\right)^2 \varb[\beta_k]\left[\frac{\partial}{\partial \beta}\log\pstrb(\vv) \right]\cnteq
 &+K\sum_{k=0}^K \delta(\beta_k)O(\dbeta^3), \nonumber\label{approx_detail:eq}
\end{align}
The second term of the r.h.s. vanishes if $K\rightarrow\infty$ as $\left|K\sum_{k=0}^K \delta(\beta_k)O(\dbeta^3)\right| \leq CK\sum_{k=0}^K \delta(\beta_k)\left|\dbeta^3\right| < C\tilde{C}^3 K\sum_{k=0} \delta(\beta_k) \left|\tk[k+1]-\tk\right|^3 = O(K^{-1}) \rightarrow 0$ with $\exists\tilde{C}, C>0$.
Note that we have $\left|\beta_{k+1} - \beta_k\right| < \tilde{C}\left|\tk[k+1]-\tk\right|$ because $\beta(t)(\in{\cal C}^2)\in{\cal C}^1$ and $|\delta(\beta)|<\infty$ because $p_\beta$ is smooth. The scaled variance is dominated by the first term of the r.h.s., which have the following limit as $K\rightarrow\infty$
\begin{align}
 \obj \triangleq \int_0^1\vbeta^2 \varb\left[\frac{\partial}{\partial \beta}\log\pstrb(\vv)\right] \dd t.
\end{align}
Therefore, $K\var\left[\log\wais\right]\rightarrow \obj$. 
\end{proof}

\section{Derivation of \refeq{de:eq}}
Euler-Lagrange equation for $\obj$ is 
\begin{align}
 \frac{\dd}{\dd t}\left(\frac{\partial G}{\partial \vbeta}\right) = \frac{\partial G}{\partial \beta}, \label{EL:eq}
\end{align}
where $G\triangleq \vbeta^2 g(\beta)$. The l.h.s. is computed as $\frac{\dd}{\dd t}(2\vbeta g(\beta)) = 2(\abeta g(\beta) + \frac{\dd g}{\dd \beta}\vbeta^2)$. The r.h.s. is computed as $\frac{\dd g}{\dd \beta}\vbeta^2$. By replacing both sides of \refeq{EL:eq} with these results, we have $\abeta + \frac{1}{2g}\frac{\dd g}{\dd \beta}\vbeta^2=\abeta + \frac{\vbeta^2}{2}\frac{\dd}{\dd \beta}\log g(\beta)= 0$

\end{document}